\documentclass{article}

\usepackage{arxiv}

\usepackage[utf8]{inputenc} % allow utf-8 input
\usepackage[T1]{fontenc}    % use 8-bit T1 fonts
\usepackage{hyperref}       % hyperlinks
\usepackage{url}            % simple URL typesetting
\usepackage{booktabs}       % professional-quality tables
\usepackage{amsfonts}       % blackboard math symbols
\usepackage{nicefrac}       % compact symbols for 1/2, etc.
\usepackage{microtype}      % microtypography
\usepackage{lipsum}		% Can be removed after putting your text content
\usepackage{graphicx}
\usepackage{natbib}
\usepackage{doi}
\usepackage{latexsym}
\usepackage{times}
\usepackage{xcolor}
\usepackage{longtable}
\usepackage{multirow}
\usepackage{subcaption}

\title{The Choice of Knowledge Base in Automated Claim Checking}

\author{ \href{https://orcid.org/0000-0003-1631-3020}{\includegraphics[scale=0.06]{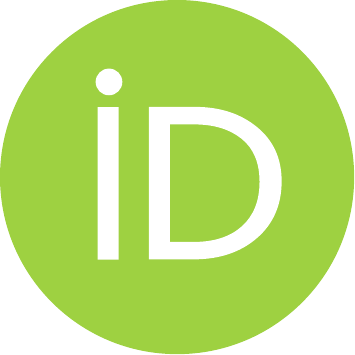}\hspace{1mm}Dominik Stammbach} \\
	Center for Law \& Economics \\
	ETH Zurich \\
	\texttt{dominsta@ethz.ch} \\
	\And
	\href{https://orcid.org/0000-0002-4439-8212}{\includegraphics[scale=0.06]{orcid.pdf}\hspace{1mm}	Boya Zhang} \\
	Department of Radiology \\ and Medical Informatics \\
	University of Geneva \\
	\texttt{boya.zhang@unige.ch} \\
	\And 	\href{https://orcid.org/0000-0002-6817-7529}{\includegraphics[scale=0.06]{orcid.pdf}\hspace{1mm}	Elliott Ash} \\
 	Center for Law \& Economics \\
 	ETH Zurich \\
 	\texttt{ashe@ethz.ch} \\
}

\renewcommand{\shorttitle}{Choice of KB in Automated Claim Checking}

\hypersetup{
pdftitle={The Choice of Knowledge Base in Automated Claim Checking},
pdfsubject={q-bio.NC, q-bio.QM},
pdfauthor={Dominik Stammbach, Boya Zhang, Elliott Ash},
pdfkeywords={Automated Claim Verification, Knowledge Bases, Evidence Selection, Information Retrieval, Content Management},
}

\begin{document}
\maketitle

\begin{abstract}
Automated claim checking is the task of determining the veracity of a claim given evidence found in a knowledge base of trustworthy facts. While previous work has taken the knowledge base as given and optimized the claim-checking pipeline, we take the opposite approach -- taking the pipeline as given, we explore the choice of the knowledge base. Our first insight is that a claim-checking pipeline can be transferred to a new domain of claims with access to a knowledge base from the new domain. Second, we do not find a \textit{"universally best"} knowledge base -- higher domain overlap of a task dataset and a knowledge base tends to produce better label accuracy. Third, combining multiple knowledge bases does not tend to improve performance beyond using the closest-domain knowledge base. Finally, we show that the claim-checking pipeline's confidence score for selecting evidence can be used to assess whether a knowledge base will perform well for a new set of claims, even in the absence of ground-truth labels. 
\end{abstract}

% keywords can be removed

\section{Introduction}

Facing a steady stream of mis-and disinformation promulgated by digital-age technologies \cite{vosughi}, a new strand of research investigates the (partial) automation of claim checking using Natural Language Processing (NLP) methods \cite[e.g.][]{vlachos_riedel, fever_dataset, content_managament_perspective}. The datasets underlying automated claim-checking systems generally have two components. First, there is a set of annotated claims, which can for example be true, false, or non-verifiable. Second, there is a knowledge base (KB) or fact base, comprising a corpus of dependable documents which can be queried and consulted to assess the veracity of the claims.\footnote{This use of the term \textit{knowledge base} is different from \textit{knowledge graphs} consisting of semantic triplets (subject-predicate-object triplets) used in other domains in NLP.}  From this KB, we retrieve the required evidence, which in turn leads to the veracity prediction for a claim. 

\begin{table*}[ht!]
    \centering
    \footnotesize
    \caption{Example Claims, with Verification using Evidence from Different Knowledge Bases}
    \label{tab:climate-fever-qualitative-analysis}

    \begin{tabular}{p{2cm} p{6.4cm} p{6.4cm}} \toprule 
    Claim & Wikipedia Evidence & Evidence from Scientific Abstracts \\ \hline
    Global warming is driving polar bears toward extinction. \textbf{True label: \textsc{Supported}} & Habitat destruction: {\color{blue}Rising global temperatures}, caused by the greenhouse effect, {\color{blue}contribute to habitat destruction}, endangering various species, such as the {\color{blue}polar bear}. Polar bear: "Bear hunting caught in global warming debate". Global warming: Rising temperatures push bees to their physiological limits, and could cause the extinction of bee populations. Extinction risk from global warming: "Recent Research Shows Human Activity Driving Earth Towards Global Extinction Event". \textbf{predicted label: \textsc{Not Enough Info}} & \textcolor{blue}{Polar bears will largely disappear} from the southern portions of their range by mid-century (Stirling and Derocher, 2012). While the {\color{blue}polar bear is the most well-known species imperiled by global warming}, and the first to be listed under the ESA solely due to this factor, it was not the first species protected under the statute in which global warming played a significant role. This highly publicized milestone firmly cemented the {\color{blue}polar bear as the iconic example of the devastating impacts of global warming} on the planet's biodiversity (Cummings and Siegel, 2009). \textbf{predicted label: \textsc{Supported}} \\ 
    The main greenhouse gas is water vapour \textbf{True label: \textsc{Supported}} & Greenhouse gas: "AGU Water Vapor in the Climate System". Global warming: As {\color{blue}water is a potent greenhouse gas, this further heats the climate}: the water vapour feedback. Global warming: The main reinforcing feedbacks are the water vapour feedback, the ice–albedo feedback, and probably the net effect of clouds. \textbf{predicted label: \textsc{Not Enough Info}} & \textcolor{blue}{Water vapour is the most abundant and powerful greenhouse gas} in Earth's atmosphere, and is emitted by human activities (Sheerwood et al., 2018). {\color{blue}Water vapour is a key greenhouse gas} in the Earth climate system (Trent et al., 2018). \textbf{predicted label: \textsc{Supported}}
  \\ 
    \end{tabular}
\end{table*}

In the previous literature on textual claim verification, the claims and the KB have been seen as part of a single unified whole. For example, FEVER consists of synthetic claims derived from Wikipedia sentences and uses Wikipedia itself as the KB \cite{fever_dataset}. Another dataset, SciFact, considers scientific claims and evidence from a selected sample of scientific paper abstracts \cite{scifact}. The task in these datasets then consists of finding the right evidence from the respective KB and to predict the correct veracity of the claim.

% add this somewhere else: \footnote{In non-textual claim verification, e.g. using knowledge graphs as in, the question which knowledge graphs to consider plays a similar critical role. Similarly, if we were to take a SQL-based approach, the question remains from which table to select the information, which is similar as selecting the appropriate KBs}

%\cite{ferreira-vlachos-2016-emergent}  \cite{fakenewschallenge} probably later

This self-contained approach to automated claim-checking certainly simplifies the scope of the problem. But it imposes a significant constraint on the practical usefulness and transferability of the resulting systems. By design, these systems will be specialized in the specific domain and less effective for checking diverse real-world claims \cite[e.g.][]{bekoulis}. Consistent with this idea, some recent work has shown that a claim-checking system that queries Wikipedia performs poorly at checking claims from a climate-change-focused task \citep{diggelmann2021climatefever}, from scientific paper abstracts \citep{scifact}, or even from general-knowledge claims that are phrased more journalistically or colloquially \citep{thorne2021evidencebased,kim-etal-2021-robust}. 

Thus, the most recent work has recognized a need for closer attention on the choice of knowledge base in automated claim checking \cite[e.g.][]{content_managament_perspective,claim_review_climate_science}. This recognition arises in parallel with stunning NLP breakthroughs brought  by self-supervised learning applied to massive, curated corpora, hinting at potential gains from a more data-expansive approach \citep[e.g.][]{bert,gpt2,gpt3}. As has been recently explored in other fields  \cite{mlops}, it is likely that automated claim checking could benefit from a more \textit{data-centric} approach, which seeks performance gains via data work, rather than the standard model-centric approach, which focuses on model architectures \cite{mlops}.

This paper puts the data-centric perspective into practice. In the standard approach, the data is held constant, and the claim-checking system is developed to improve performance on existing benchmarks \cite[see e.g.][]{fever_dataset, scifact, augenstein2019multifc, kotonya2020explainable}. We do the opposite: The claim-checking system is held constant, while we systematically permute the KB corpus and the dataset of checked claims.  Thus we generalize the process of automated claim-checking by detaching the claims from the KB. 

Table \ref{tab:climate-fever-qualitative-analysis} illustrates the core intuition, with two climate-science-related claims listed in Column 1. A claim-checking system using a KB of Wikipedia articles does not produce sufficient evidence to resolve the claim (Column 2). When using a KB of scientific abstracts, however, the same claim-checking system recovers stronger supporting evidence and can reproduce the ground-truth label prediction (Column 3). Put simply, a scientific claim requires a scientific KB for a claim-checking system to predict the ground-truth label. This paper explores the generality of this insight. 

Our experimental setup works as follows. We take a standard claim-checking system that performs evidence retrieval followed by natural language inference between the evidence and the claim \citep{e-fever}. We then consider a battery of claim verification tasks and knowledge bases. For each <claim set, KB> pair, we apply the pipeline without any further training.\footnote{A similar approach is taken on a smaller scale by \cite{claim_review_climate_science}, who retrieved evidence for Climate FEVER from Wikipedia and a subset of PubMed.}

We perform claim-checking on six labeled claim tasks. Besides two tasks based on Wikipedia \citep{fever,foolmetwice} and one on scientific paper abstracts \citep{scifact}, we have a task on climate-change-related claims \citep{diggelmann2021climatefever}, one on statements from the 2016 presidential debates \citep{clef_2019}, and one on real-world information needs based on search-engine queries \citep{thorne2021evidencebased}. Our paper is the first to take on all of these diverse claim verification tasks using a single system. In turn, we confront these diverse verification tasks with knowledge bases from diverse domains. We use KB's on general knowledge (the universe of Wikipedia article summaries), from the scientific domain (a scientific KB consisting of more than 70Mio scientific abstracts), and from the journalistic domain (the universe of recent New York Times articles). In addition, we experiment with combining these knowledge bases, further generalizing the effort by \citet{fakta} that combines Wikipedia with a corpus of online news articles. Finally, we compare them to "the whole internet" queried via a search engine \citep[e.g.][]{popat_2016, karadzhov-etal-2017-fully, augenstein2019multifc, clef_2019}. 

In the experiments, we observe large differences in performance for claims from the various datasets using different knowledge bases. In particular, we show that a claim-checking pipeline can obtain good performance in a new domain (e.g., from Wikipedia to science articles) without additional model training, as long as it has access to a KB from the new domain. In line with Table \ref{tab:climate-fever-qualitative-analysis}, intuitively, we obtain the highest label accuracy for each claim task using the KB that most closely matches the domain of the claims. In general, lower performance is driven by failure to retrieve the required evidence and by assigning support/refute predictions to non-verifiable claims, rather than making incorrect support/refute determinations.  

Meanwhile, the union of multiple knowledge bases tends to perform similarly to the single KB from the closest domain. Sometimes, the combined KB performs worse. Thus, the issue of choosing the most appropriate KB cannot be easily resolved by pooling all available knowledge bases. To address this issue, we ask whether the evidence pipeline metrics are predictive of system performance. We show that the BM25 similarity between a claim and the closest KB documents performs poorly as a metric. However, the confidence score produced by the evidence selection module (a fine-tuned RoBERTa-based classifier) performs well. Thus, this evidence quality metric could be used to forecast system performance in new domains without labels. We demonstrate the usefulness of data-driven KB selection by showing that the overall best system across all claim-verification tasks is to select the KB with the highest evidence quality score by task.

To summarize, this paper makes the following contributions:

\begin{enumerate}
\item We take a claim verification pipeline as given and explore its behaviour on different claim-checking tasks.
\item We perform experiments where we hold the pipeline constant, but vary the knowledge base from which evidence is retrieved.
\item We observe large gains in label accuracy if we use a more suitable knowledge base for a given claim-checking task.
\item We find that combining multiple KBs does not automatically yield better results. We investigate alternative methods to choose the most suitable KB for a given claim-checking task.
\end{enumerate}

These findings highlight the pivotal role of the knowledge base for automated claim checking. We provide additional evidentiary support for previous papers making this and related points \cite{content_managament_perspective,claim_review_climate_science, goasdou_et_al}.  We hope these results motivate the enhancement of existing knowledge bases and production of new ones. Further, it could be that additional performance gains are achievable through joint optimization of the KB and the claim-checking system. 

The insights from this paper might be applicable in other settings besides text-based claim verification. For example, the benefits from focusing on unstructured knowledge bases might also pay out in the case of structured knowledge graphs \citep{wilcke2017knowledge}, leading to improvements in non-textual claim verification systems \citep{tchechmedjiev2019claimskg, tien_duc_cao, scrutinizer}. Meanwhile,  open-domain question answering tasks have much in common with automated claim verification \cite{zhu2021retrieving}. One can easily imagine similar performance gains from selecting KB's according to the question type before searching for answers.

\section{Methods}\label{sec:data}

\subsection{Overview}

    \begin{figure}[!h]
        \caption{Overview of the Claim-checking Task}
        \label{fig:system_detail}
        \centering
        \includegraphics[scale=0.6]{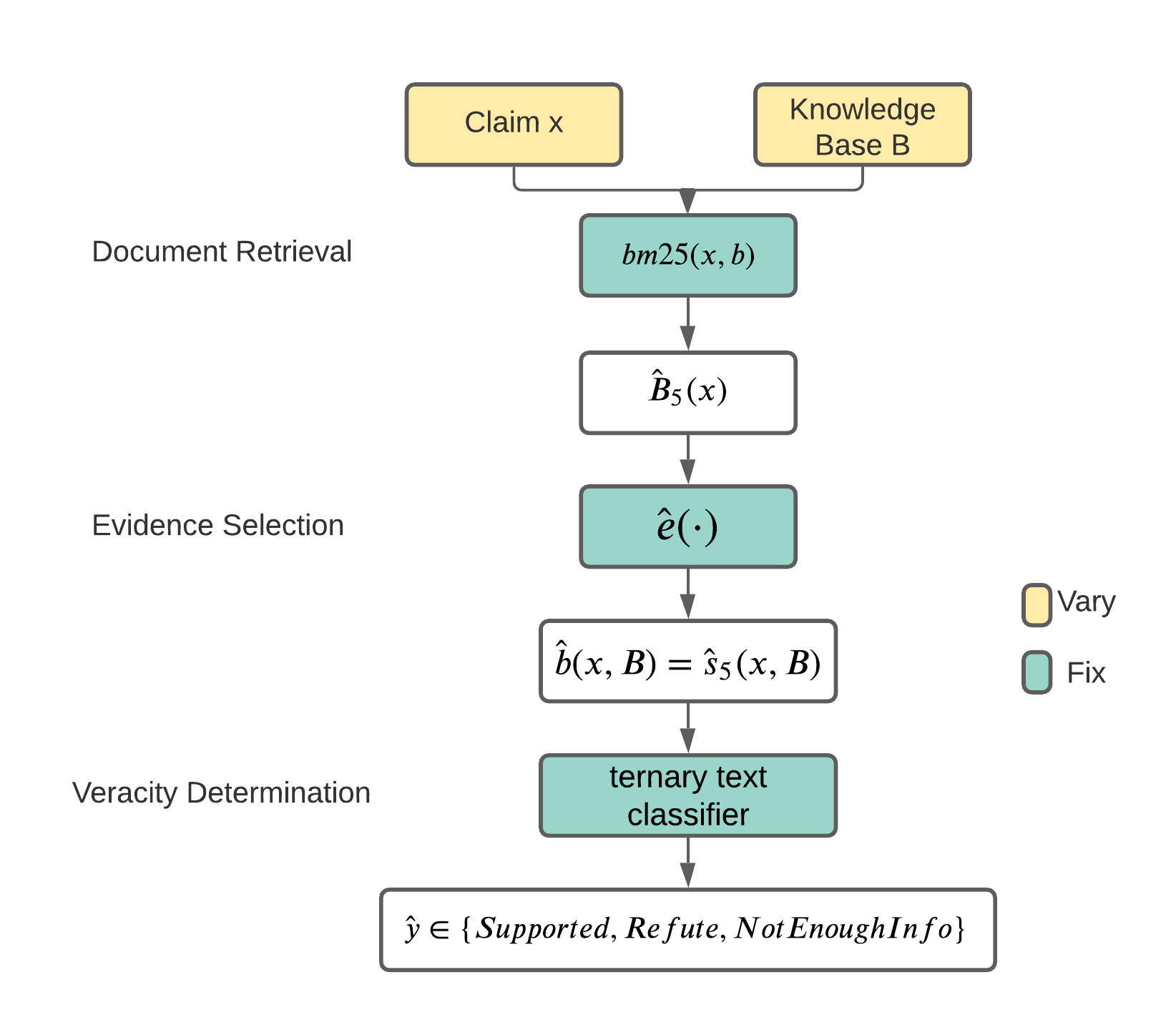}
        %\caption{Pipeline Detail}
    \end{figure}

The claim-checking task can be summarized as follows. We have a set of claims $C=(X, Y)$, consisting of a plain-text claim $x$ and a three-class veracity label, $y\in$\{\textsc{Supported}, \textsc{Refuted}, \textsc{Not Enough Info}\}. In addition, we have access to a knowledge base (KB) $B$, comprising a list of plain-text evidence statements indexed by $b$. The objective of a claim-checking system is to take as input a claim $x$ and the KB $B$ and learn a prediction function $\hat{y}(x,B)$.

Due to computational constraints, the system cannot take as input all text in $B$, so we split the problem into two steps. First, a retrieval step identifies a set of candidate evidence statements $\hat{b}(x,B)$ that serve as a sufficient statistic for the KB $B$ in the determination of claim $x$. Then the veracity prediction function is learned using the retrieved evidence, $\hat{y}(x,\hat{b}(x,B))$.

%We show an overview of our method in Figure

We show an illustration of our claim-checking approach in Figure \ref{fig:system_detail}. As we can see, the task decomposes in a data part and a modelling part. Previous work has mainly focused on optimizing $\hat{b}(\cdot)$ and $\hat{y}(\cdot)$ for a given task and associated KB, i.e., allocated resources to optimize the models holding the data fixed. Our approach is different. We hold the system ($\hat{b}(\cdot)$,  $\hat{y}(\cdot)$) fixed and vary the inputs ($C,B$). 

\begin{figure}[!h]
    \caption{Overview of Experiment Setup}
    \label{fig:experimental_setup}
    \centering
    \includegraphics[scale=0.11]{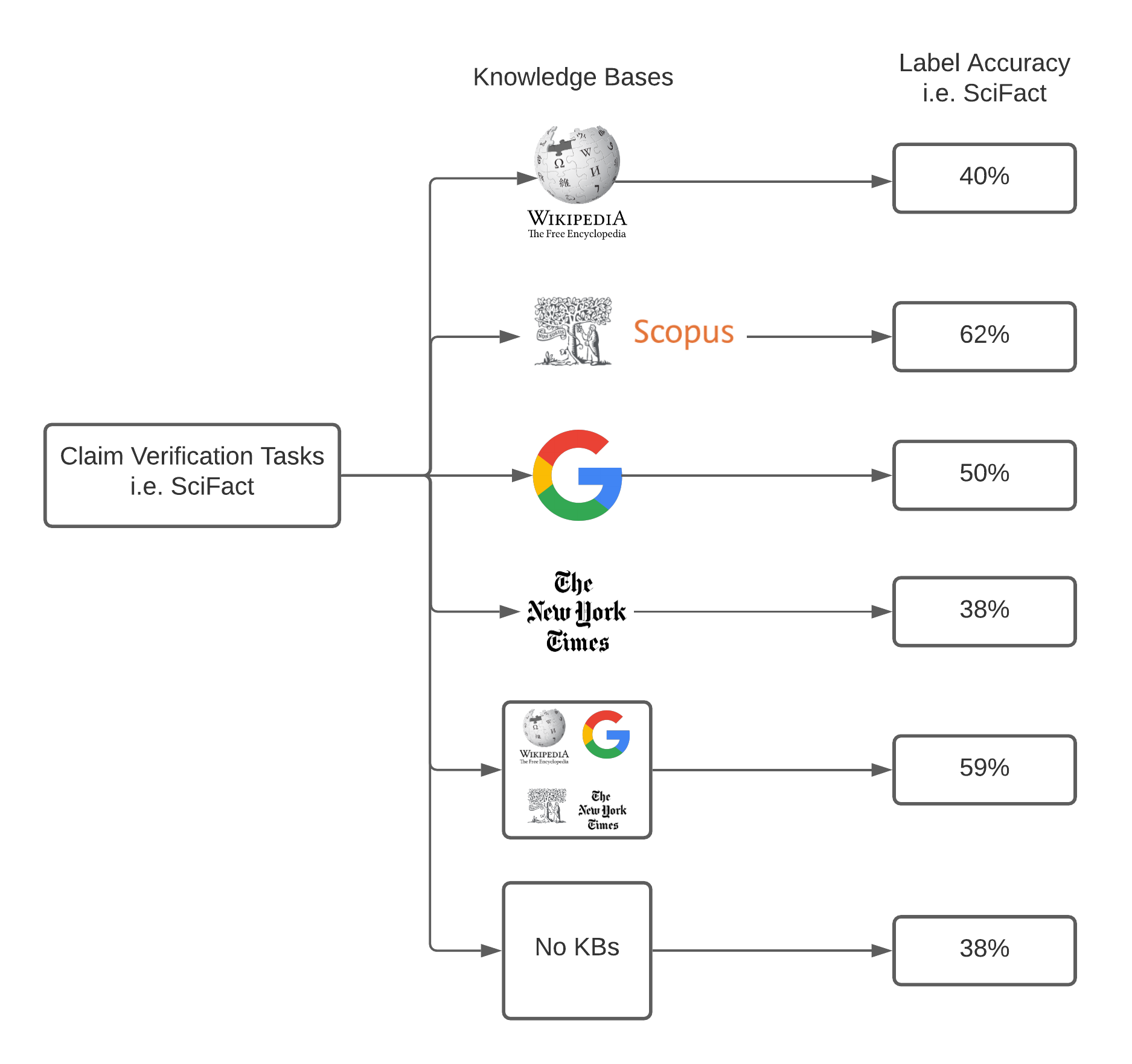}
\end{figure}

Going forward, we can refer to a particular claim task as $C_i$, indexed by $i$, and a particular KB as $B_j$, indexed by $j$. Figure \ref{fig:experimental_setup} provides an illustration, taking as an example $C_i$=SciFact. For a given scientific claim $x$, we retrieve relevant evidence  $\hat{b}(x,B_j))$ from the respective knowledge base $B_j$. Given a claim and retrieved evidence, the model predicts a claim veracity $\hat{y}(x,\hat{b}(x,B_j))$, which can be compared to the true value $y$. This process is repeated and average label accuracy is reported for each <claim set, KB> pair ($C_i,B_j$).

The rest of this section provides additional detail on this procedure. Respectively, we discuss the claim verification pipeline (2.2), the claim verification tasks (2.3), and the sourced knowledge bases (2.4).

\subsection{Claim Verification Pipeline}

Our claim verification pipeline ($\hat{b}(\cdot)$,  $\hat{y}(\cdot)$) is held fixed across experiments. The starting point is \citet{e-fever}, a system initially designed for claim-checking FEVER's synthetic claims using Wikipedia as the KB ($C_i$=FEVER, $B_j$=Wikipedia). Our only modification consists of adjusting the document retrieval step to work for arbitrary inputs ($C_i$,$B_j$). 

Our choice for using the \citet{e-fever} system is twofold. First, it is the best performing FEVER system for which we found the code and models available.\footnote{We used the code provided here: \url{https://github.com/dominiksinsaarland/domlin_fever}.} Second, the approach is close to the original FEVER baseline system's (augmented with more powerful Transformer architectures), and thus a sensible standard choice when the focus is on the knowledge base. Nevertheless, our results are not contingent on using the \citet{e-fever} system. In Appendix Table \ref{tab:results_zero_shot_fever_kgat}, we can replicate our main results using another publicly available FEVER system \cite{kgat}.\footnote{Our main experiments, the additional experiments listed in the Appendix, and the ones conducted in \cite{claim_review_climate_science} indicate that the knowledge base play a crucial role in evidence-based textual claim verification, regardless of the pipeline. However, we leave it to future work to further investigate whether this insight generalizes in all cases for all pipelines.}

The first step is retrieval of relevant documents. For a given claim $x$, all documents $b$ in $B$ are ranked by BM25($x$,$b$), the BM25 similarity between $b$ and $x$ \citep{anserini}.\footnote{This document retrieval step is the  major difference with \citet{e-fever}, which was too specialized for Wikipedia -- in particular, using the MediaWiki API and following hyperlinks.}  The top-five documents on this ranking, denoted as $\hat{B}_5(x)$, are retrieved as containing potential evidence.

The second step is selection of evidence statements from the retrieved documents. After splitting the sentences using spaCy \citep{spacy2}, each sentence $s \in \hat{B}_5(x)$ is assigned a score $\hat{e}(s,x)\in(0,1)$, interpretable as the probability that $s$ provides evidence about the claim $x$. The evidence scoring function $\hat{e}(\cdot)$, borrowed directly from \citet{e-fever}, is a RoBERTa-based binary text classifier trained to identify evidence statements in an annotated Wikipedia-based dataset. Let $\hat{s}_5(x,B)$ be the top-five sentences ranked by evidence score, across the sentences in all retrieved documents $\hat{B}_5$. These sentences provide the evidence supplied to the veracity classifier. That is, $\hat{b}(x,B)=\hat{s}_5(x,B)$. 

The third step of the pipeline is to make a veracity determination based on the claim $x$ and extracted evidence $\hat{s}_5(x,B)$. This module consists of a ternary text classifier assigning predicted probabilities across the classes $\hat{y}\in$\{\textsc{Supported}, \textsc{Refuted}, \textsc{Not Enough Info}\}. Again borrowed directly from \citet{e-fever}, the veracity classifier is a fine-tuned RoBERTa model using FEVER's annotated dataset of claims and evidence from Wikipedia. For training and inference, the claim statement and the evidence statements are concatenated as a single string.

\subsection{Claim Verification Tasks}

For the set of claims $C=(X,Y)$, we examine six different tasks. For each task, we evaluate all claims in the official development set, if present. We include more detailed descriptions in Appendix \ref{appendix:claims}, for example on how the labels were configured to make them comparable across tasks. Summary statistics on these tasks -- e.g. number of examples, class distributions -- are reported in Appendix Table \ref{tab:dataset_descriptions}. Appendix Table \ref{tab:example_claims1} shows a sample of data points for each task. 

\begin{enumerate} 
    \itemsep2pt 
    \item \textbf{FEVER} is a large-scale task based on Wikipedia sentences, and then manually verified by a second set of annotators using Wikipedia who did not know the source of the claim \citep{fever}. We use the first 2K claims from the development set.\footnote{One of our examined KBs is the Google API. Due to rate limits of 100 queries a day, it was not possible to experiment on all 19'998 claims from the FEVER development set.}
    \item \textbf{SciFact} SciFact considers the task of scientific claim verification \cite{scifact}. The claims are generated from sentences in citing articles by annotators with expertise in scientific NLP and life science, based on a corpus of 5,183 scientific abstracts that have been sampled from Semantic Scholar \citep{semanticscholar}. We ran our experiments on the 300 claims from the development set. 
    \item \textbf{Climate FEVER} contains new annotated claims related to climate change, with a mixture of journalistic and scientific claims \citep{diggelmann2021climatefever}. Because of the lack of a development set in this task, we considered all 1,381 claims which were not disputed in our experiments. 
    \item \textbf{Presidential Speeches} is a task covering two presidential debates and one vice presidential debate held in the 2016 U.S. elections \cite{clef_2019}. It contains 70 annotated true or false claims on which we base our analysis. We excluded half-true sentences and sentences without any annotation. 
    \item \textbf{Real-World Information Needs} is a task with annotated claims based on search engine queries, representing real-world information needs \cite{thorne2021evidencebased}. We include the development set of this task. Similar to climate FEVER, we produce an aggregated label for each claim and exclude claims with conflicting annotations, resulting in 930 claims.
    \item \textbf{Fool Me Twice} is a task containing challenging entailment pairs gathered through a multi-player game based on Wikipedia \cite{foolmetwice}. The aim of the game is to construct claims which are difficult to verify. We use the 1,169 claims from the development set.
\end{enumerate}

% \begin{figure}[!tp]
%     \caption{Overview of Experiment Setup}
%     \label{fig:experimental_setup}
%     \centering
%     \includegraphics[scale=0.13]{figures/experimental_setup.png}
% \end{figure}

\subsection{Knowledge Bases} 

The component of interest in our study is the knowledge base (KB) $B$. Each KB, described in this subsection, is a set of plain-text documents. Appendix Table \ref{tab:stats_knowledge_bases} reports summary statistics on each KB, including the number of documents and the average document length. Appendix Table \ref{tab:examples_knowledge_bases} shows a sample of sentences from each KB. In the following, we describe the four knowledge bases in more detail.

\paragraph{General-Knowledge Domain: Wikipedia.}
The first KB is a corpus of the introductory summaries for all English-language Wikipedia articles (N = 5.5 million) as of 2017. This corpus was used to construct the FEVER claim-checking task \citep{fever_dataset}.\footnote{Because this dump was popularized in FEVER, we decided to use it as is, opposed to e.g., newer versions of Wikipedia}  It is a general-knowledge crowd-sourced encyclopedia emphasizing breadth rather then depth. 

\paragraph{Scientific Domain: Scientific Abstracts.}

The scientific KB is built using a large corpus of scientific abstracts. These abstracts are compiled from three sources. The bulk of the abstracts (61Mio abstracts) come from Scopus, a large-scale database managed by Elsevier. Another 8Mio abstracts are added from CrossRef. The last 8Mio abstracts come from the Semantic Scholar Project \cite{semanticscholar}. In total, the KB covers 77Mio abstracts from scientific articles in all scientific fields. The papers from which we collected the abstracts were published between 1990-2020.

\paragraph{Journalistic Domain: N.Y. Times.}

The third KB is from the news-media domain, as investigated by %Nadeem et al.
\cite{fakta, ferreira-vlachos-2016-emergent, fakenewschallenge}. For this purpose, we use all full-text articles crawled from the New York Times web site, published between January 2000 and March 2021 (N = 2 million).

\paragraph{"The Whole Internet": Google Search API.}
We have a fourth KB representing the statements available on the searchable internet. This is arguably the broadest possible domain. For this KB, our document retrieval system works differently. We use the Google Search API to retrieve documents from the web. Each query is the plain claim, for which we retrieve 10 hits from the API. We use these as the retrieved documents $\hat{B}_5$.\footnote{Arguably, giving the Google API 10 hits rather than 5 hits is not a fair comparison with the other databases. We did this because the snippets are very short. We also produced our main experiment results limiting to just the five most relevant snippets and the results were nearly identical.} After that, the pipeline proceeds normally with the evidence statements selected from the preview snippets.

\paragraph{All KB's.}

Beside the four individual knowledge bases, we analyze a synthetic KB that includes the union of the four domains. For each claim, the top-five documents in each KB (so 20 documents in total) are included as potential evidence. Then the top 5 sentences by the score $\hat{e}$, across all 20 documents, are used as evidence.

\paragraph{No KB.}

As a minimal baseline, we produce a prediction using just the claim as input without considering any evidence. By construction, this setting almost always produces the label \textsc{Not Enough Info}.

\paragraph{"Best Evidence" KB.}

We use an unsupervised measure to automatically choose the most appropriate KB for a given claim verification task. This measure will be discussed in more detail in Section \ref{sec:evidence_quality}.

\section{Results}

\begin{table*}[ht!]
\caption{Label Accuracy Across Claim Verification Tasks Using Different Knowledge Bases}
  \label{tab:results_zero_shot_fever}
\centering
\renewcommand{\arraystretch}{1.5}
\begin{tabular}{l | c c c c  c c | c}
 \hline    
 &  \multicolumn{6}{c}{\textbf{Claim Verification Task}} \\ 
 \hline
 \textbf{Knowledge Base} & FEVER & SciFact & Climate & Presidential & Real-World & Fool Me Twice & Avg.\\ \hline
Wikipedia & \textbf{74} & 39 & \underline{43} & 1 & \underline{38} & \underline{24} & 37 \\
Science Abstracts & 47 & \textbf{60} & \textbf{45} & 1 & 31 & 9 & 32 \\
NYTimes & 55 & 39 & 43 & \underline{10} & 36 & 16 & 33 \\
Google API & \underline{72} & \underline{50} & 36 & \textbf{26} & \textbf{61} & \textbf{40}  & 47 \\
%Google API & \underline{72} & \underline{49} & 36 & \textbf{23} & \textbf{62} & \textbf{39}  \\ % only top5 hits google API

\hline
None & 38 & 38 & 34 & 0 & 14 & 0 & 21 \\ 
All & 75 & 59 & 47 & 21 & 54 & 41 & 50 \\
Best Evidence & 74 & 60 & 43 & 26 & 61 & 40 & 51 \\
\hline

\end{tabular}
\flushleft \small{\textit{Notes}. Development set label accuracy of different claim-checking tasks (columns) using different KB domains (rows). Best results for single KB experiments by task/column in bold, second-best underlined. In the last column, we show average scores of a KB on all experiments. In the bottom half of the table, we show results of synthetic experiments.}
\end{table*}

\subsection{Main Results}

The results of our main analysis are reported in Table \ref{tab:results_zero_shot_fever}. Each cell reports the experiment's label accuracy for the column-indicated task using the row-indicated knowledge base (KB). The last column provides the averaged results for a given KB on the different claim verification tasks. 

The third-to-last last row corresponds to the minimal baseline inputting the claim but without any evidence (and therefore mechanically predicting \textsc{Not Enough Info}). The penultimate row shows results using the union of all knowledge bases, to be discussed in Subsection \ref{sec:all-KB}. The last row indicates results using an unsupervised measure to automatically detect the most appropriate KB, to be discussed in Subsection \ref{sec:evidence_quality}. 

In the rest of this subsection, we provide a detailed discussion by claim verification task (that is, by column). Throughout the discussion, we refer to Appendix Table \ref{tab:conf-mat-all}, which displays full confusion matrices for the results of each experiment. Appendix Table \ref{tab:bootstrapped_main_results} provides results with bootstrapped confidence intervals.

\paragraph{FEVER.}

First, it is not surprising that the best label accuracy for the FEVER task (74\%) was achieved using Wikipedia. The FEVER task consists of synthetic claims based on Wikipedia articles, and our claim-checking pipeline was originally designed to work on FEVER by consulting Wikipedia articles.\footnote{Notably, the change to BM25 for document retrieval obtains comparable results to the original system in \citet{e-fever}} Thus, it is also not surprising that this experiment gets the highest accuracy across all single-KB experiments.

The second-best results are obtained using the Google Search API (72\%). After manual investigation of some examples, we find that the API often returns the required Wikipedia page to verify a claim. Meanwhile, the NYT and Science KB's perform significantly worse. Consulting the confusion matrices (Appendix Table \ref{tab:conf-mat-all}), we see that with both of these KB's, the pipeline almost always predicts \textsc{Not Enough Info}. Intuitively, with these out-of-domain knowledge bases, the pipeline cannot recover enough useful evidence to check the claims.

\paragraph{SciFact.}

The second task, SciFact, consists of scientific claims. As the Science Abstracts KB comprises abstracts from scientific articles, it is not surprising that this KB delivers the best label accuracy. At 60\% accuracy, the Science KB is significantly and proportionally better than the second-best Google API at 50\%.

Stepping back for a moment, the strong level of performance (60\%) is itself notable. That performance level is comparable to baseline systems which were trained on SciFact, notwithstanding that our pipeline has never before seen the SciFact claims task data. On this task, as we will see again in the other tasks, our system that has only seen FEVER claims can still deliver decent performance. 

However, the new KB is a necessary ingredient in that performance. Using the pipeline's native KB, Wikipedia, delivers poor performance on SciFact. As with the NYT KB, performance with Wikipedia is barely better than the no-evidence baseline.  As can be seen with the confusion matrices, these out-of-domain KB's predict almost exclusively the third class due to lack of evidence.

\paragraph{Climate-FEVER.}
Climate-FEVER contains a mix of journalistic and scientific claims about climate change. Reflecting the relevance of the scientific claims, the best label accuracy (45\%) is achieved using the Scientific KB. However, the difference is not nearly as stark as with SciFact: Wikipedia (43\%) and NYTimes (43\%) are not that much worse. The confusion matrices suggest that none of the KB's are well-suited to this task, because all of them generate way too many \textsc{Not Enough Info}'s.  Interestingly, this is the only task where the Google API returns the worst results.  %The KB used in the original paper was Wikipedia which provided our second best experimental results. 

\paragraph{Presidential Debates.}
In the presidential debates task, we obtain best results using the Google Search API -- yet still quite low at 26\%. The NYT KB is worse at 10\%,\footnote{This result illustrates the over-all usefulness of the New York Times KB, which performs consistently better than a claim-only baseline and on par with Wikipedia for the experiments which are not based on Wikipedia claims. We expect such a journalistic KB would be even better if bigger and more carefully curated (for example removing quotes).} while the other databases are close to zero. This lower performance likely reflects that the style of speaking in presidential debates -- political language -- is far away from the encyclopedic, scientific, and journalistic domains.  In addition, the lower average than the other categories reflects that in this task, there are no examples of \textsc{Not Enough Info}. In the confusion matrix, we find that barely any of the false claims could be identified across all experiments, while with some evidence from e.g. Google, we managed to verify some of the true statements.

\paragraph{Real-World Claims.}

For the real-world claims task, we achieve by far the best label accuracy again using the Google Search API -- a respectable 61\%. This is proportionally quite a lot better than the other knowledge bases, including the second-placed Wikipedia (38\%). One interpretation of this result is that the writing style of the claims, plain English, is unlike the more specialized language in the three main KB's which might result in BM25 retrieval failure. Google Search works well given its access to many plain-English web sites. As illustrated with the confusion matrices, these differences are not as extreme as with the specialized tasks, and each KB can do better than the minimal baseline.

\paragraph{Fool Me Twice.}

The KB originally associated with the task again is Wikipedia, and we find the same trend observed in the Real-World Claims task. Best results come from the Google Search API, followed by the other three knowledge bases (with Wikipedia performing second-best). As seen with the confusion matrices, the results are also driven by lack of retrieved evidence. The evidence retrieval system trained with FEVER cannot yet deal with the challenging problem formulations in Fool Me Twice. \\

%\vspace{2pt}

\noindent Overall, we find mixed results in these experiments. We conclude that there is no \textit{"universally best"} knowledge base. We find the most broad KB, the Google API, often is a reasonable default, but is outperformed by more suitable KBs on half of the tasks examined. Hence, one should be careful in selecting the KB for automated claim checking systems. 

Moreover, we have an indication of where and why the systems tend to fail. Appendix Table  \ref{tab:wiki-conf-mat} shows a single aggregated confusion matrix that sums across all the single-KB tasks. That matrix illustrates that the \textsc{Supported} and \textsc{Refuted} classes have high precision. If one looks just at the top-left $2 \times 2$ matrix, one would infer a highly performant system, as it is rare for the model to make an incorrect decision between \textsc{Supported} and \textsc{Refuted}. Moreover, it is also rare for the model to mistakenly assign a veracity value (\textsc{Supported} or \textsc{Refuted}) when the true value is \textsc{Not Enough Info}. The bulk of the mistakes made by the model consist of incorrectly predicting the \textsc{Not Enough Info} label.  

This tendency is reassuring as it suggests the system to be somewhat "modest" in its classifications. If the proper facts to verify a claim are missing in a KB, our system at least predicts that claims then are not verifiable. Conversely, high label accuracy corresponds to the facts actually being present in a KB.

\subsection{Combining Knowledge Bases}
\label{sec:all-KB}

For each claim-checking task, we run an additional experiment considering the top-five document hits from all the individual KBs. This modified system gives the evidence selection module access to additional documents. Since all of the useful evidence from the individual KB's is available, one might expect that combining databases would only improve performance. 

As shown in Table \ref{tab:results_zero_shot_fever} (penultimate row), however, this is not the case. The combined KB is similar in performance to the best single KB. With FEVER, Climate-FEVER, and Fool Me Twice, the combined KB is only slightly better than any single KB.\footnote{For e.g. FEVER, we found in manual inspections that the addition of the Google API corrected some of the mistakes from BM25 using just Wikipedia by yielding the required Wikipedia page.} %For Climate-FEVER ..., for FM2 ...}
This tiny increase in performance is perhaps most surprising for Climate-FEVER, where the claims come from both a scientific and journalistic domain. In that context, especially, one might expect benefits from combining scientific and journalistic knowledge bases. However, that expectation is not borne out in the experiments. 

\begin{figure}[hbt!]
    
    \caption{Evidence Retrieval Errors when Combining Knowledge Bases}
    \label{fig:error_analysis}
    
    \centering
    \scriptsize
    \begin{tabular}{p{6cm}} 
\textbf{(a) Claim:}: Autologous transplantation of mesenchymal stem cells causes a higher rate of opportunistic infections than induction therapy with anti-interleukin-2 receptor antibodies. \\ \textit{[claim derived from \citet{scifact_devset_error_analysis}]} \\
\textbf{Highest Scoring Evidence From Science KB}: Among patients undergoing renal transplant, the use of autologous MSCs compared with anti-IL-2 receptor antibody induction therapy resulted in lower incidence of acute rejection, decreased risk of opportunistic infection, and better estimated renal function at 1 year \\ \textit{[evidence from \citet{scifact_devset_error_analysis}, in Science Abstracts KB]} \\
\textbf{Highest Scoring Evidence from All Knowledge  Bases} [...] In high-risk kidney transplant recipients, induction therapy with rabbit but is associated with significant toxicity, opportunistic infections, and cancer. Using reduced doses of RATG combined with anti–IL-2 antibodies may achieve the mAb against the CD25 subunit of the IL-2 receptor (IL-2R) on activated T cells, such as Tocilizumab, a humanized monoclonal antibody against the IL-6 receptor already used in Anti-TNF-naïve patients had higher remission and response rates than safety profile without any risk of serious or opportunistic infections [147] [...] \\ \textit{[evidence retrieved from \citet{google_evidence_1} and \citet{google_evidence_2} using Google API]} 
    \end{tabular}
\end{figure}

Meanwhile, the combined KB actually sometimes produces more errors than the best single KB. Figure \ref{fig:error_analysis} shows an example of such an error in the SciFact task. For this claim, with access only to the Science KB produces the correct evidence and the system correctly predicts \textsc{Refuted}. But with access to all knowledge bases, the system has access to additional evidence from the Google API which rank highly on the evidence scoring function yet turn out not to be useful to the veracity classifier. Hence, the All-KB system guesses incorrectly, \textsc{Not Enough Info}.

In some cases, then, adding more potential evidence documents can reduce performance by reducing the quality of the selected evidence. One possible implication is that the issues of KB choice cannot be solved by pooling all knowledge bases. Another possibility is that evidence retrieval and evidence selection systems need to be adapted to accommodate larger and more diverse knowledge bases. This is an important area for further investigation.

\subsection{Evidence Quality}\label{sec:evidence_quality}

The results reported so far suggest that the quality of the retrieved and selected evidence  is pivotal for the functioning of a claim-checking pipeline. Motivated by this notion, we investigate this directly using our system outputs as metrics for evidence quality. 

An appealing interpretation of our main results is that if we have a high topical overlap between a set of claims and a knowledge base (KB), we would tend to retrieve the appropriate facts and obtain a higher label accuracy. A direct measure of this overlap is provided by the BM25 scores of the outputs in the document retrieval step. Thus, to test this interpretation of the results, we compare the average label accuracy for a given claims-KB pair with the associated average BM25 score for the highest-ranked retrieved document.

\begin{figure}
\centering
\caption{Evidence Quality and Label Accuracy}
\label{fig:corr-evidence-la}

\begin{subfigure}[b]{0.55\textwidth}

   \caption{A. Accuracy vs. Retrieved Document Similarity (max BM25)}
   \includegraphics[scale=0.5]{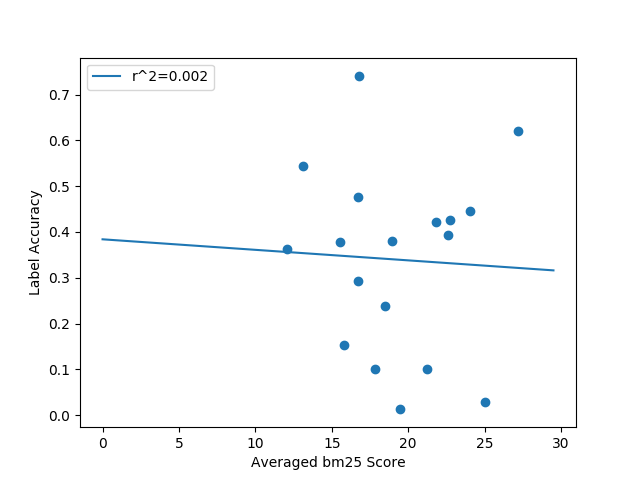}
   \end{subfigure}

\begin{subfigure}[b]{0.55\textwidth}
   \caption{B. Accuracy vs. Selected Evidence Score (max $\hat{e}$)}
   \includegraphics[scale=0.5]{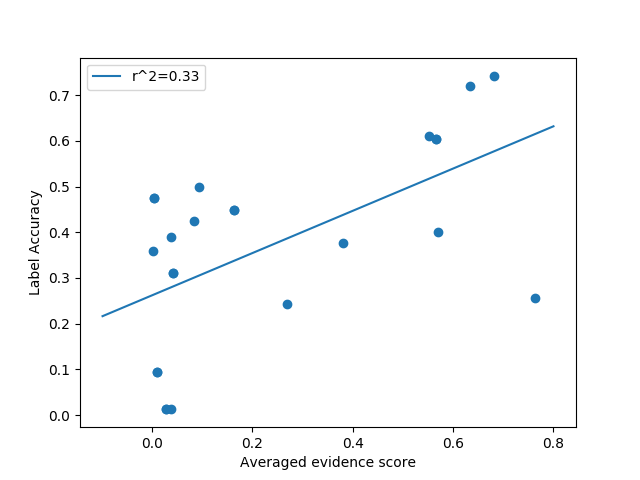}
\end{subfigure}
\end{figure}

% \begin{figure}
%  \caption{Evidence Quality and Label Accuracy}
%     \label{fig:corr-evidence-la}
%     \centering
%     \small{A. Accuracy vs. Retrieved Document Similarity (max BM25)}
%     \includegraphics[scale=0.45]{figures/scatter_bm25.png} 
%     \small{B. Accuracy vs. Selected Evidence Score (max $\hat{e}$)} 
%     \includegraphics[scale=0.45]{figures/scatterplot_evidence_veracity.png}
% \end{figure}

Figure \ref{fig:corr-evidence-la} Panel A shows a scatterplot of this relationship. Each datapoint corresponds to a <claims, KB> pair, with the x-axis giving the average of the highest BM25 score across claims and the y-axis giving the label accuracy from the associated experiment. We find that BM25 scores of retrieved documents are not correlated with the resulting label accuracy. The estimated Pearson correlation is -0.05and not statistically significant ($p=0.85$).  Appendix Table \ref{tab:bm25_scores}, reporting the numbers by task-KB pair, shows the same story. For example, the Science KB tends to obtain a high top-BM25 score across all tasks, even though it is not the best KB for most tasks. Thus, the intuition that a standard text similarity metric like BM25 would identify a useful KB turns out to be a wrong one.  

Given that BM25 similarity does not capture knowledge-base suitability, we assess a second potential pipeline metric -- the confidence score $\hat{e}$ provided by the evidence selection module. Figure \ref{fig:corr-evidence-la} Panel B is complementary with Panel A, plotting average label accuracy by claims-KB pair against the average of the highest assigned evidence score across claims. Here, there is a clear positive relationship. 
%The slope coefficient is 0.3885, e.g. for each 0.1 increase in evidence selection confidence of our model, the resulting label accuracy increases by roughly 4\% points. 
The statistically significant relationship has an $r^2$ of $.33$ and a Pearson correlation of 0.49 ($p=0.015$). The numbers in Appendix Table \ref{tab:evidence_scores} illustrate how each task's same-domain KB, which tends to produce the highest accuracy, also obtain the highest evidence score. 

What is the practical usefulness of this relationship between evidence scores and label accuracy? To check for this, we ran an additional synthetic experiment where, for each claim task, the KB with the highest average evidence confidence score is selected.  The results are reported in the last row of Table \ref{tab:results_zero_shot_fever}. We find that this data-driven KB-choice strategy matches the best individual KB's label accuracy in five out of six tasks. Further, the approach obtains the highest average accuracy across all tasks (rightmost column), slightly better than the union of all KB's.

As a complementary experiment, we performed KB selection at the claim level (rather than the task level). That is, for each claim, we fetch evidence from each KB and then use the KB with the best evidence quality score. The claim-level best-evidence approach did slightly better on half the tasks, but worse on the other half, and overall worse on average (Appendix Table \ref{tab:bootstrapped_main_results}, bottom row). Looking deeper into the predictions, we found that the lower performance is driven in part by the model being \textit{too} confident in the produced evidence. It tends to make veracity determinations even when the correct label is \textsc{Not Enough Info} (Appendix Table \ref{tab:conf-mat-all}, bottom row).

In summary, high overlap of surface-form TF-IDF features does not indicate the suitability of a KB and a set of claims. However, the confidence score produced by the sentence retrieval module provides a metric for the suitability of a KB for checking a set of claims. While this is what the evidence selection module is designed to do, it is nonetheless reassuring that it works in practice. It illustrates the additional, and pivotal, semantic information extracted by the RoBERTa-based model which is not captured by surface TF-IDF features. 

These results could have significant practical value. The confidence metric can be used to provide an immediate indication whether a claim-checking pipeline, along with a particular KB, can be transferred to checking a new domain of unlabeled claims.  Given the costs of producing and labeling true and false claims, this approach makes fact-checking systems more useful by providing an indication of their transferability. Further, in new domains where one has multiple KB's to choose from, the evidence quality metric provides guidance on which KB to apply.

\section{Conclusion}

This work has explored the choice of the knowledge base (KB) in automated claim checking. We used a single system to predict label accuracy for a number of claim-checking tasks from different domains while varying the KB and holding everything else constant. We have shown that FEVER-style claim-verification systems have capacity for checking out-of-domain claims, as long as the system has access to facts from the new domain. Overall, choosing a suitable KB for such out-of-domain claims matters a lot and we can improve the resulting label accuracy by a large margin using a more suitable KB. We also find that a larger KB is not always better -- e.g., combining all knowledge bases often does not result in better label accuracy than taking the most suitable one. 

Our approach and findings are in line with a resurgent data-centric paradigm in machine learning. This paradigm takes the view that the data used for a machine learning task is at least as important as the model. While in some ways this is an old idea, for example in the context of feature engineering \citep[e.g.][]{zheng2018feature}, the recognition of the importance of data quality has gained renewed attention in deep learning due to the impressive gains in large-scale language modeling made through curation of datasets \citep[see e.g.][]{gpt2, gpt3, t5}. In particular, \citet{thepile} achieve remarkable improvements in language modeling using more compact architectures but with a more carefully curated pre-training corpus. Given the diminishing returns to more complex architectures, increasing data quality still leaves much room for improvement.

These insights have clear relevance for automated claim verification. While efforts to create more diverse lists of checkable claims are of course valuable, a more balanced approach would investigate all data dependencies, among them the choice of the knowledge bases used for claim-checking. Our experiments show that with a given FEVER-based system, zero-shot label accuracy can increase by as much as 20\% with a more appropriate KB (e.g. with SciFact) compared to the standard choice of using evidence retrieved from Wikipedia. In light of these results, researchers in automated claim verification may decide to curate their knowledge bases as much as their pipeline architectures.  

\bibliographystyle{unsrtnat}
%\bibliographystyle{apalike}
%\bibliography{references}
\bibliography{bibliography}

\appendix

\setcounter{figure}{0} \renewcommand{\thefigure}{A.\arabic{figure}} \setcounter{table}{0} \renewcommand{\thetable}{A.\arabic{table}}

\newpage

\section{Additional Information on Claims Tasks}
\label{appendix:claims}

\paragraph{Claim Task Data Sources.}
The following items include some additional information on the claims task datasets.

\begin{enumerate} 
    \itemsep2pt 
    \item \textbf{FEVER} The whole dataset consists of 185,445 claims, split in a train set/development set and (blind) testset. Because we can only query 100 claims per day using the Google Search API, we restricted our experiment to the first 2K FEVER claims from the development set. We ran our experiments for three knowledge bases (Wikipedia, Scientific KB and NYT KB) for all 19'998 claims in the development set and found that results are nearly identical to the first 2K claims. 
    \item \textbf{SciFact} SciFact contains in total 1.4K annotated claims, split in train/development and (blind) testset. We use the official development set in our experiments. 
    \item \textbf{Climate FEVER} The dataset consists of 1.5K claims. Following the evaluation procedure of the baseline system in the original paper, we (1) use their aggregated label for each claim and (2) exclude disputed claims in our analysis.
    \item \textbf{Presidential Speeches} is a dataset covering two presidential debates and one vice presidential debate held in the 2016 U.S. elections. In total, it covers 4064 sentences, however the vast majority are not annotated. The dataset contains contains 70 annotated claims which are either true or false. We use these 70 claims in our experiments. An additional 24 claims are annotated as half-true which we excluded.
    \item \textbf{Real-World Information Needs} contains 11K claims in total, split in train/development and test set. Annotation is performed sentence-wise. Similar to Climate FEVER, we have chosen 930 examples from the development set in which there are no conflicting veracity labels, i.e. one evidence sentence supports a claim whereas another refutes the claim. We excluded 158 such conflicting claims. 
    \item \textbf{Fool Me Twice} contains 13K challenging entailment pairs gathered through a multi-player game based on Wikipedia. It is split in train/dev and test set and we use the development set in our analysis. 
\end{enumerate}

\paragraph{Claim Task Label Processing.}

We had to reconfigure some of the labels in the verification tasks to make them comparable. First, some of the tasks recommend not to use aggregated label accuracy. Climate-FEVER includes an aggregated label per claim only to measure the performance of a FEVER system on their dataset. In the Real-World Claims dataset, each claim-evidence pair has a unique annotation. We aggregate the labels in the same way as in Climate-FEVER and exclude claims with conflicting veracity scores. Next, veracity labels in SciFact are associated with rationales -- i.e., the combined evidence needed to support or refute a claim. We found that there were no conflicting rationales for any claim in the dataset, however, so this part is ignored. Lastly, we are aware that FEVER Score is determined not only by predicting the right label, but also retrieving the right evidence. Because we are testing out new unlabeled knowledge bases, we do not explore this part of FEVER.

\paragraph{Claim Task Summary Statistics.}

In Table \ref{tab:dataset_descriptions}, we provide summary statistics on the subset of claims we examined from the datasets present in our main analysis. 

\begin{table}[ht!]
\centering
\caption{Dataset sizes and Detailed Class Distributions}
\label{tab:dataset_descriptions}
\renewcommand{\arraystretch}{1.5}
\scalebox{0.8}{\begin{tabular}{l |c c c c c}
\hline
Dataset & Claim Domain & N & \textsc{Supported} (\%) & \textsc{Refuted} (\%) & \textsc{Not Enough Info} (\%) \\ \hline
FEVER & Wikipedia & 2000 & 33.3\% & 33.3\% & 33.3\% \\ 
SciFact & Scientific & 300 & 41.3\% & 21.3\% & 37.3\%\\ 
Climate-FEVER & Scientific & 1381 & 47.4\% & 18.3\% & 34.3\% \\ 
Presidential & Political & 70 & 31.4\% & 68.6\% & 0\% \\ 
Real-World & Search Engine & 930 & 57.5\% & 29.3\% & 13.2\% \\ 
Fool Me Twice & Wikipedia & 1169 & 51\% & 49\% & 0\% \\ \hline
%8,723 1,088 1,086
\end{tabular}}
\end{table}

\paragraph{List of Example Claims.}

Table \ref{tab:example_claims1} shows example claims with labels from each of the task datasets used in the main analysis.

\begin{table}[ht!]
\centering
\caption{Examples for Datasets}
  \label{tab:example_claims1}
\renewcommand{\arraystretch}{1}
\begin{longtable}{l l p{10cm}}
\hline
\textbf{Dataset} & \textbf{Example} \\ \hline
\multirow{6}{*}{\textbf{FEVER}} & \textbf{Claim:} &Fox 2000 Pictures released the film Soul Food. \\
& \textbf{Label:} & \textsc{Supported} \\  
& \textbf{Claim:} &Savages was exclusively a German film. \\
& \textbf{Label:} & \textsc{Refuted} \\ 
& \textbf{Claim:} &Ashley Cole plays the tuba. \\
& \textbf{Label:} & \textsc{Not Enough Info} \\ \hline 
\multirow{6}{*}{\textbf{SciFact}} & \textbf{Claim:} &1/2000 in UK have abnormal PrP positivity. \\
& \textbf{Label:} & \textsc{Supported} \\
& \textbf{Claim:} &A total of 1,000 people in the UK are asymptomatic carriers of vCJD infection. \\
& \textbf{Label:} & \textsc{Refuted} \\
& \textbf{Claim:} &0-dimensional biomaterials show inductive properties. \\
& \textbf{Label:} & \textsc{Not Enough Info} \\ \hline
\multirow{6}{*}{\textbf{Climate-FEVER}} & \textbf{Claim:} &The Great Barrier Reef is experiencing the most widespread bleaching ever recorded. \\
& \textbf{Label:} & \textsc{Supported} \\ 
& \textbf{Claim:} &The polar bear population has been growing. \\
& \textbf{Label:} & \textsc{Refuted} \\ 
& \textbf{Claim:} &Burping cows are more damaging to the climate than all the cars on this planet. \\
& \textbf{Label:} & \textsc{Not Enough Info} \\ \hline 
\multirow{4}{*}{\textbf{Presidential}} & \textbf{Claim:} &The state of Indiana has balanced budgets. \\
& \textbf{Label:} & \textsc{Supported} \\  
& \textbf{Claim:} &Independent analysts say the Clinton plan would grow the economy by 10.5 million jobs. \\
& \textbf{Label:} & \textsc{Refuted} \\ \hline
\multirow{6}{*}{\textbf{Real-World}} & \textbf{Claim:} &Queen margaery dies in game of thrones. \\
& \textbf{Label:} & \textsc{Supported} \\
& \textbf{Claim:} &The limbic system is part of the cerebral cortex. \\
& \textbf{Label:} & \textsc{Refuted} \\
& \textbf{Claim:} &Michelin tires are the same as michelin stars. \\
& \textbf{Label:} & \textsc{Not Enough Info} \\\hline
\multirow{4}{*}{\textbf{Fool Me Twice}} & \textbf{Claim:} & Filming for the movie Gandhi in India was delayed due to political unrest. \\
& \textbf{Label:} & \textsc{Supported} \\ 
& \textbf{Claim:} & Raging Bull is based  on a true event about an out of control bull. \\
& \textbf{Label:} & \textsc{Refuted} \\ \hline
\end{longtable}
\end{table}
%in the appendix, list of data point examples from all claims and all knowledge bases 

\clearpage

\section{Additional Material on Knowledge Bases}
\label{appendix:KB}

\paragraph{Knowledge Base Summary Statistics.}

Table \ref{tab:stats_knowledge_bases} provides summary statistics on the knowledge bases used in the main analysis. 

\begin{table}[ht!]
\centering
\caption{Knowledge Bases Statistics Summary}
  \label{tab:stats_knowledge_bases}

\renewcommand{\arraystretch}{1}
\begin{tabular}{l | c c c}
\hline
Knowledge Base & N documents & Average N Sentences & Average N words \\ \hline
Wikipedia & 5.4M & 5 & 102 \\
Scientific KB  & 76.5M & 8 & 178  \\ 
NYTimes & 1.9M & 70 & 1474  \\ \hline 
\end{tabular}
\end{table}

\paragraph{List of Knowledge Base Evidence Statements.}

Table \ref{tab:examples_knowledge_bases} shows example sentences sampled from the knowledge bases used in the main analysis.

\begin{table}[ht!]
\centering
\caption{Examples for Knowledge Bases}
  \label{tab:examples_knowledge_bases}

\renewcommand{\arraystretch}{1}
\begin{tabular}{l | p{10cm}}
\hline
Knowledge Base & Example of A Part of Contents\\ \hline
Wikipedia & Since 1980 , a significant global warming has led to glacier retreat becoming increasingly rapid and ubiquitous , so much so that some glaciers have disappeared altogether , and the existences of many of the remaining glaciers are threatened . The acceleration of the rate of retreat since 1995 of key outlet glaciers of the Greenland and West Antarctic ice sheets may foreshadow a rise in sea level , which would affect coastal regions .\\ \hline
Scientific KB  & Sea surface temperature (SST) is the interface between the ocean and the overlying atmosphere. Estimating sea surface temperature from infrared satellite and in situ temperature data Sea surface temperature (SST) is a critical quantity in the study of both the ocean and the atmosphere as it is directly related to and often dictates the exchanges of heat, momentum, and gases between the ocean and the atmosphere.   \\  \hline
NYTimes & The real concern is whether further warming and shrinkage of sea ice might drive the bears to extinction. In a wrenching scene in “An Inconvenient Truth“, the documentary that propelled a wave of interest in human-caused global warming in 2006, a polar bear slowly slips beneath the waves to drown.  \\   \hline
Google Search API & Climate change includes both global warming driven by human-induced emissions of These clouds reflect solar radiation more efficiently than clouds with fewer and While aerosols typically limit global warming by reflecting sunlight, black The effects of climate change on humans, mostly due to warming and shifts in Oct 31, 2011 Dark roofs absorb sunlight that heats up your house, office tower, more heat- trapping greenhouse gas emissions, and more global warming. Cooler buildings need less air conditioning, which translates to fewer roof in the fight against climate change: cover it with solar panels. \\ \hline 
\end{tabular}
\end{table}

\clearpage

\section{Supporting Experiment Results}
\label{appendix:results}

\begin{table*}[ht!]
\caption{Label Accuracy Across Claim Verification Tasks Using Different Knowledge Bases using KGAT}
  \label{tab:results_zero_shot_fever_kgat}
\centering
\renewcommand{\arraystretch}{1.5}
\begin{tabular}{l | c c c c  c c }
 \hline    
 &  \multicolumn{6}{c}{\textbf{Claim Verification Task}} \\ 
 \textbf{Knowledge Base} & FEVER & SciFact & Climate & Presidential & Real-World & Fool Me Twice \\ \hline
Wikipedia & \textbf{64} & 41 & 43 & 16 & \underline{44} & \textbf{40}  \\
Science Abstracts & 52 & \textbf{46} & \textbf{44} & 6 & 30 & 16  \\
NYTimes & \underline{57} & 40 & \underline{44} & \underline{26} & 41 & 28  \\
Google API & 34 & \underline{46} &
35 & \textbf{31} & \textbf{47} & \underline{36} \\
\hline
\end{tabular}
\flushleft \small{\textit{Notes}. This table is equivalent to our main results, using a different Claim Verification System, namely KGAT \cite{kgat}. We show development set label accuracy of different claim-checking tasks (columns) using different KB domains (rows). Best results by task/column in bold, second-best underlined (for the single KB experiments). Spearman's Rank Correlation of the results obtained by KGAT and our main results is 67.7, which is highly significant with $p < 0.0005$}, indicating that our experimental findings translate similarly to other FEVER systems. 

\end{table*}

These tables show confusion matrices to provide additional information on how the predictions compare with the true classes. Rows indicate true labels, while columns indicate model predictions. The rows and columns correspond to \textsc{Supported}, \textsc{Refuted}, and \textsc{Not Enough Info}, respectively. To assist interpretation, confusion matrix values are normalized such that the cells add up to 100. 

Table \ref{tab:wiki-conf-mat} shows the aggregated matrix across all single-KB tasks. Table \ref{tab:conf-mat-all} shows a confusion matrix for each task-KB pair. In Table \ref{tab:conf-mat-all} bottom row (Oracle), we see the true label distributions for each task.

In the No KB row, we see that the model (almost) always guesses "not enough evidence". A third-column prediction means that the KB does not contain sufficient evidence or that the model has not found it. 

\begin{table}[ht!]
    \centering
        \caption{The averaged Confusion Matrix for the single-KB tasks}
    \label{tab:wiki-conf-mat}

    \begin{tabular}{c |c c c c}
    \hline  
         &  \textsc{Supported} & \textsc{Refuted} & \textsc{Not Enough Info} \\ \hline
        %\textsc{Supported} & 28 & 1 &  5 \\ 
        %\textsc{Refuted} & 2 & 22 & 9 \\
        %\textsc{Not Enough Info} & 6 & 4 & 24 \\
        
        \textsc{Supported} & 16 & 1 & 27 \\
        \textsc{Refuted} & 2 & 10 & 20 \\
        \textsc{Not Enough Info} & 3 & 2 & 19 \\
        \hline

    \end{tabular}
    
    \flushleft Table sums across all the matrices in the first four rows of Table \ref{tab:conf-mat-all}. Rows are normalized frequencies per class, while columns are normalized predictions per class.
    %\caption{The FEVER + Wikipedia Confusion Matrix (rows are normalized frequencies per class, columns are normalized predictions per class)}
\end{table}

\begin{table*}[ht!]
\caption{Bootstrapped Label Accuracy and 95\% Confidence Intervals}
  \label{tab:bootstrapped_main_results}
\centering
\renewcommand{\arraystretch}{1.5}
\begin{tabular}{l | c c c c  c c }
 \hline    
 &  \multicolumn{6}{c}{\textbf{Claim Verification Task}} \\ 
 \textbf{Knowledge Base} & FEVER & SciFact & Climate & Presidential & Real-World & Fool Me Twice \\ \hline

wiki & 74.21 +- 1.95 & 39.24 +- 5.36 & 42.6 +- 2.72 & 1.34 +- 2.77 & 37.72 +- 3.15 & 24.4 +- 2.32 \\
science & 47.31 +- 2.04 & 60.35 +- 5.53 & 45.04 +- 2.5 & 1.61 +- 2.99 & 31.2 +- 3.05 & 9.43 +- 1.63 \\
nyt & 54.92 +- 2.3 & 38.72 +- 5.48 & 43.01 +- 2.53 & 9.94 +- 6.86 & 36.08 +- 3.02 & 15.73 +- 2.03 \\
googleAPI & 71.96 +- 1.97 & 49.92 +- 5.48 & 36.01 +- 2.57 & 25.58 +- 10.17 & 61.2 +- 3.14 & 40.32 +- 2.69 \\ \hline
None & 37.89 +- 2.04 & 37.7 +- 6.19 & 34.51 +- 2.55 & 0.0 +- 0.0 & 14.22 +- 2.45 & 0.44 +- 0.35 \\
All & 75.33 +- 1.95 & 59.36 +- 5.77 & 46.61 +- 2.51 & 20.98 +- 9.53 & 54.14 +- 2.99 & 41.18 +- 2.76 \\
Best Evidence (Task) & 74.19 +- 1.87 & 60.22 +- 5.37 & 42.9 +- 2.74 & 26.14 +- 9.7 & 61.12 +- 2.93 & 40.08 +- 2.88 \\
Best Evidence (Claim) & 75.17 +- 1.75 & 60.72 +- 5.35 & 46.2 +- 2.62 & 18.25 +- 9.74 & 50.14 +- 3.38 & 39.11 +- 2.86 \\
\hline
\end{tabular}
\flushleft \small{\textit{Notes}. Development set label accuracy of different claim-checking tasks (columns) using different KB domains (rows). We report the mean of bootstrapped sampling with N=200 and report 95\% confidence intervals (+- 1.96 * standard error). }
\end{table*}

\begin{table}
\centering
\caption{All Confusion Matrices for all Experiments}
\label{tab:conf-mat-all}

\begin{tabular}{|l |  c   c   c|  c   c   c |  c   c   c |  c   c   c |  c  c   c |  c   c   c |}
\hline    
 &  \multicolumn{18}{c|}{\textbf{Fact Checking Task}} \\ 
 \hline
 \textbf{KB} & \multicolumn{3}{c}{FEVER} & \multicolumn{3}{c}{SciFact} &  \multicolumn{3}{c}{Climate} & \multicolumn{3}{c}{President.} & \multicolumn{3}{c}{Real-W.} & \multicolumn{3}{c|}{Fool Twice} \\ \hline

\multirow{3}{*}{Wikipedia} & 28 & 1 & 5 & 3 & 1 & 37 & 5 & 1 & 42 & 0 & 0 & 31 & 21 & 3 & 33 & 14 & 3 & 34 \\
 & 2 & 22 & 9 & 1 & 3 & 18 & 1 & 5 & 12 & 4 & 1 & 63 & 4 & 8 & 17 & 3 & 10 & 36 \\
 & 6 & 4 & 24 & 3 & 1 & 33 & 1 & 1 & 32 & 0 & 0 & 0 & 3 & 2 & 9 & 0 & 0 & 0 \\ \hline
\multirow{3}{*}{Science} & 8 & 1 & 25 & 26 & 2 & 13 & 9 & 1 & 37 & 1 & 0 & 30 & 14 & 2 & 41 & 5 & 1 & 44 \\
 & 1 & 9 & 23 & 2 & 12 & 8 & 1 & 5 & 12 & 4 & 0 & 64 & 1 & 5 & 23 & 1 & 4 & 44 \\
 & 1 & 1 & 31 & 10 & 5 & 22 & 2 & 2 & 31 & 0 & 0 & 0 & 1 & 1 & 12 & 0 & 0 & 0 \\ \hline
\multirow{3}{*}{NYT} & 15 & 1 & 18 & 3 & 1 & 38 & 10 & 1 & 36 & 10 & 1 & 20 & 18 & 4 & 35 & 10 & 2 & 40 \\
 & 1 & 13 & 19 & 0 & 2 & 19 & 2 & 5 & 12 & 27 & 0 & 41 & 2 & 7 & 20 & 2 & 6 & 41 \\
 & 3 & 2 & 28 & 3 & 1 & 34 & 4 & 2 & 28 & 0 & 0 & 0 & 2 & 1 & 11 & 0 & 0 & 0 \\ \hline
\multirow{3}{*}{Google API} & 30 & 1 & 3 & 21 & 1 & 20 & 1 & 0 & 46 & 26 & 0 & 6 & 44 & 5 & 9 & 28 & 1 & 22 \\
 & 2 & 20 & 11 & 3 & 6 & 12 & 1 & 1 & 17 & 49 & 0 & 20 & 8 & 13 & 8 & 5 & 12 & 32 \\
 & 8 & 3 & 22 & 11 & 3 & 23 & 0 & 1 & 34 & 0 & 0 & 0 & 5 & 5 & 4 & 0 & 0 & 0 \\ \hline
\multirow{3}{*}{No KBs} & 1 & 0 & 33 & 0 & 0 & 41 & 0 & 0 & 47 & 0 & 0 & 31 & 1 & 0 & 56 & 0 & 0 & 51 \\
 & 0 & 4 & 28 & 0 & 1 & 21 & 0 & 0 & 18 & 0 & 0 & 69 & 0 & 0 & 28 & 0 & 0 & 49 \\
 & 0 & 0 & 33 & 0 & 0 & 37 & 0 & 0 & 34 & 0 & 0 & 0 & 0 & 0 & 13 & 0 & 0 & 0 \\ \hline
\multirow{3}{*}{All KBs} & 31 & 1 & 3 & 28 & 1 & 12 & 14 & 1 & 32 & 21 & 1 & 9 & 38 & 5 & 15 & 28 & 2 & 21 \\
 & 2 & 23 & 7 & 1 & 12 & 8 & 3 & 5 & 10 & 47 & 0 & 21 & 7 & 11 & 11 & 6 & 13 & 30 \\
 & 7 & 4 & 22 & 13 & 6 & 19 & 5 & 2 & 27 & 0 & 0 & 0 & 5 & 4 & 5 & 0 & 0 & 0 \\ \hline
\multirow{3}{*}{Best Evidence (Task)} & 28 & 1 & 5 & 26 & 2 & 13 & 10 & 1 & 36 & 26 & 0 & 6 & 44 & 5 & 9 & 28 & 1 & 22 \\
 & 2 & 22 & 9 & 2 & 12 & 8 & 2 & 5 & 12 & 49 & 0 & 20 & 8 & 13 & 8 & 5 & 12 & 32 \\
 & 6 & 4 & 24 & 10 & 5 & 22 & 4 & 2 & 28 & 0 & 0 & 0 & 5 & 5 & 4 & 0 & 0 & 0 \\ \hline
\multirow{3}{*}{Best Evidence (Claim)} & 31 & 1 & 2 & 28 & 1 & 11 & 14 & 1 & 33 & 19 & 0 & 13 & 33 & 5 & 20 & 27 & 1 & 23 \\ 
 & 2 & 23 & 8 & 3 & 12 & 6 & 3 & 5 & 11 & 46 & 0 & 23 & 7 & 11 & 11 & 5 & 13 & 31 \\
 & 8 & 4 & 21 & 13 & 5 & 19 & 5 & 2 & 28 & 0 & 0 & 0 & 4 & 3 & 6 & 0 & 0 & 0 \\ \hline
 Oracle & 34 & 32 & 33 & 41 & 21 & 37 & 47 & 18 & 34 & 31 & 69 & 0 & 56 & 28 & 13 & 51 & 49 & 0 \\ \hline
\end{tabular}
\flushleft \small{\textit{Notes}. Rows are true values, while columns give predicted values.  Within each 3x3 matrix, values are normalized to sum to 100. Rows/columns respectively correspond to \textsc{Supported}, \textsc{Refuted}, \textsc{Not Enough Info}. See Table \ref{tab:wiki-conf-mat} for a labeled confusion matrix for the averaged confusion matrix of the four individual KBs; The last row (Oracle) shows the diagonal of the confusion matrix where always the right label is predicted.} 

\end{table}
\clearpage

\begin{table}[ht!]
\caption{BM25 Scores Across Claim Verification Tasks Using Different Knowledge Bases}
  \label{tab:bm25_scores}
\centering
\renewcommand{\arraystretch}{1.5}
\begin{tabular}{l | c c c c  c c}
 \hline    
 &  \multicolumn{6}{c}{\textbf{Claim Verification Task}} \\ 
 \textbf{Knowledge Base} & FEVER & SciFact & Climate & Presidential & Real-World & Fool Me Twice \\ \hline
% these are top 5 bm25 averaged scores
%Wikipedia & 14.3 & 20.0 & 19.7 & 18.2 & 13.8 & 16.2  \\
%Science Abstracts & 15.5 & 25.2 & 22.7 & 23.5 & 15.6 & 19.6 \\
%NYTimes & 12.0 & 17.0 & 21.1 & 16.8 & 11.3 & 14.6 \\ \hline 

% these are top 1, thus numbers are slightly higher
Wikipedia & 16.7 & 22.6 & 21.8 & 19.5 & 15.5 & 18.5 \\
Science Abstracts & 16.7 & 27.2 & 24.1 & 25.0 & 16.7 & 21.2 \\
NYTimes & 13.1 & 19.0 & 22.7 & 17.8 & 12.1 & 15.8 \\ \hline 

\end{tabular}
\flushleft \small{\textit{Notes}. Development set BM25 scores of different claim-checking tasks (columns) using different knowledge base domains (rows). Google API is not included because of the different document retrieval approach (see text).}

\end{table}

\begin{table}[ht!]
\caption{$\hat{e}$ Scores Across Claim Verification Tasks Using Different Knowledge Bases}
  \label{tab:evidence_scores}
\centering
\renewcommand{\arraystretch}{1.5}
\begin{tabular}{l | c c c c  c c}
 \hline    
 &  \multicolumn{6}{c}{\textbf{Claim Verification Task}} \\ 
 \textbf{Knowledge Base} & FEVER & SciFact & Climate & Presidential & Real-World & Fool Me Twice \\ \hline
Wikipedia & 0.55 & 0.16 & 0.24 & 0.17 & 0.46 & 0.4  \\
Science Abstracts & 0.09 & 0.54 & 0.33 & 0.18 & 0.22 & 0.11 \\
NYTimes & 0.18 & 0.1 & 0.33 & 0.49 & 0.37 & 0.17 \\  
Google API & 0.57 & 0.41 & 0.02 & 0.59 & 0.55 & 0.55 \\  \hline

\end{tabular}
\flushleft \small{\textit{Notes}. Development set $\hat{e}$ scores of different claim-checking tasks (columns) using different knowledge base domains (rows)}

\end{table}

\clearpage

\section{Results on Additional Tasks}
\label{app:more-datasets}

\begin{table}[ht!]
\centering
\caption{Examples for Datasets not included in our main analysis}
  \label{tab:example_claims2}

\renewcommand{\arraystretch}{1}
\begin{tabular}{l l p{9cm}}
\hline
\textbf{Dataset} & \textbf{Example} \\ \hline
\multirow{6}{*}{\textbf{MultiFC (filtered)}} & \textbf{Claim:} & Peachtree and Pine is one of the leading sites for tuberculosis in the nation. \\
& \textbf{Label:} & \textsc{Supported}\\ 
& \textbf{Claim:} & A photograph shows a group of children in South Africa giving a meerkat a bath. \\
& \textbf{Label:} & \textsc{Refuted} \\ 
& \textbf{Claim:} & Pope Francis: All Dogs Go to Heaven. \\
& \textbf{Label:} & \textsc{Not Enough Info} \\ \hline
\multirow{6}{*}{\textbf{Snopes}} & \textbf{Claim:} &Photograph shows a tourist who died of fright after being photographed in the Sundarbans. \\
& \textbf{Label:} & \textsc{Supported} \\ \
& \textbf{Claim:} &A photograph shows a man posing with a horse he killed in Australia with a bow and arrow. \\
& \textbf{Label:} & \textsc{Refuted} \\ 
& \textbf{Claim:} &Property owners in New York City will be fined \$250,000 for using “improper pronouns” due to new transgender laws. \\
& \textbf{Label:} & \textsc{Not Enough Info} \\ \hline 
\multirow{6}{*}{\textbf{Public Health}} & \textbf{Claim:} &Raw Milk Straight from the Cow \\
& \textbf{Label:} & \textsc{Supported} \\
& \textbf{Claim:} &Germany has banned pork from school canteens because it offends Muslim 'migrants.' \\
& \textbf{Label:} & \textsc{Refuted} \\
& \textbf{Claim:} &One out of four corporations doesn't pay a nickel in (federal income) taxes. \\
& \textbf{Label:} & \textsc{Not Enough Info} \\ \hline 
\end{tabular}
\end{table}

We have run a subset of the experiments for three additional claim verification tasks. These datasets are from journalistic sources. However, we did not include these in our main results. We are under the impression that the claims in these datasets are not \textit{complete} in the sense that a fact check about these claims can be performed given the claim alone. We refer to the discussion in \citet{diggelmann2021climatefever} for a more in-depth definition of what a valid claim is. Given the examples listed in Table \ref{tab:example_claims2} for these datasets, the claims are not self-contained nor unambiguous in their textual surface form.

\begin{table}[ht!]
\centering
\caption{Development Set Label Accuracy of Datasets not included in our main analysis}
  \label{tab:results_zero_shot_fever2}

\renewcommand{\arraystretch}{1}
\begin{tabular}{l |c c c c c}
\hline
Dataset & Wikipedia & Scientific KB & NYTimes & Google Search API & None\\ \hline
MultiFC (filtered) & 45 & 42 & 43 & 29  & 44 \\ 
Snopes & 19 & 19 & 20 & 21 & 17  \\ 
Public Health & 19 & 19 & 23 & 43 & 17 \\ \hline 
\end{tabular}
\end{table}

%The claims in these datasets are crawled from fact checking websites (or snopes). The evidence provided in these datasets is extracted from the same URL the claim is crawled from. The additional information available during dataset creation thus is the URL from which the claim is extracted. This provides further contextual information and the evidence about the claim. 
We show the results in Table \ref{tab:results_zero_shot_fever2}. Overall, we did not manage to retrieve suitable evidence from any of our knowledge bases. Hence, all our knowledge bases (and the No KB baseline) yield similar results for these datasets. 

\end{document}